\documentclass[letterpaper]{article} 
\usepackage{aaai24}
\usepackage{times}  
\usepackage{helvet}  
\usepackage{courier}  
\usepackage[hyphens]{url}  
\usepackage{graphicx} 
\usepackage{natbib}  
\usepackage{caption} 
\usepackage{algorithm}
\usepackage{algorithmic}
\usepackage{subfigure}
\usepackage{mathtools}
\usepackage{amsfonts}
\usepackage{booktabs}
\usepackage{newfloat}
\usepackage{listings}
\usepackage{xcolor}

\DeclareCaptionStyle{ruled}{labelfont=normalfont,labelsep=colon,strut=off} 
\floatstyle{ruled}
\newfloat{listing}{tb}{lst}{}
\floatname{listing}{Listing}
\title{
\texttt{STAS:} Spatial-Temporal Return Decomposition for Multi-agent Reinforcement Learning
}

\author {
    Sirui Chen\textsuperscript{\rm 1}\equalcontrib,
    Zhaowei Zhang\textsuperscript{\rm 2, \rm 4}\equalcontrib,
    Yaodong Yang\textsuperscript{\rm 2}\thanks{Corresponding author},
    Yali Du\textsuperscript{\rm 3\dag} 
}

\affiliations {
    \textsuperscript{\rm 1}School of Information, Renmin University of China\\
    \textsuperscript{\rm 2}Institute for Artificial Intelligence, Peking University\\
    \textsuperscript{\rm 3}King’s College London\\
    \textsuperscript{\rm 4}National Key Laboratory of General Artificial Intelligence, BIGAI\\
    chensr16@gmail.com,
    zwzhang@stu.pku.edu.cn, 
    yali.du@kcl.ac.uk,
    yaodong.yang@pku.edu.cn	
    
}

\usepackage{bibentry}

\begin{document}

\maketitle

\begin{abstract}

Centralized Training with Decentralized Execution (CTDE) has been proven to be an effective paradigm in cooperative multi-agent reinforcement learning (MARL). One of the major challenges is credit assignment, which aims to credit agents by their contributions. 
While prior studies have shown great success, their methods typically fail to work in episodic reinforcement learning scenarios where global rewards are revealed only at the end of the episode.
They lack the functionality to model complicated relations of the delayed global reward in the temporal dimension and suffer from inefficiencies. To tackle this, we introduce Spatial-Temporal Attention with Shapley (\textbf{STAS}), a novel method that learns credit assignment in both temporal and spatial dimensions. It first decomposes the global return back to each time step, then utilizes the Shapley Value to redistribute the individual payoff from the decomposed global reward. To mitigate the computational complexity of the Shapley Value, we introduce an approximation of marginal contribution and utilize Monte Carlo sampling to estimate it. We evaluate our method on an Alice \& Bob example and MPE environments across different scenarios. Our results demonstrate that our method effectively assigns spatial-temporal credit, outperforming all state-of-the-art baselines.

\end{abstract}

\section{Introduction}

In recent years, significant progress has been made in the field of cooperative multi-agent reinforcement learning, demonstrating its potential to solve diverse problems across several domains \cite{du2023review}, such as Real-Time Strategy games~\citep{vinyals2019grandmaster}, distributed control~\citep{4445757}, and robotics~\citep{perrusquia2021multi}. However, cooperative games present a considerable challenge known as the credit assignment problem, which involves determining each agent's contribution using the centralized controller. Furthermore, real-world environments are complex and unpredictable, leading to long-term delayed and discontinuous rewards, and causing inefficiencies in the learning process. For instance, in the game of Go, the agent may not receive rewards until the outcome is determined. These nontrivial features pose significant challenges to existing methods that mainly rely on dense rewards to learn a policy that maximizes long-term returns. Additionally, the lack of dense rewards can make finding optimal solutions difficult, resulting in high variance during training and potential failure to complete the task.

Substantial progress has been made in solving the credit assignment problem, with notable approaches including QMIX ~\citep{DBLP:conf/nips/RashidFPW20}, QTRAN ~\citep{DBLP:conf/icml/SonKKHY19}, and SQDDPG ~\citep{DBLP:conf/aaai/WangZKG20}. However, these methods do not address delayed reward environments. Recently, several studies have focused on temporal return decomposition methods that aim to decompose the return on a temporal scale ~\citep{DBLP:journals/corr/abs-1905-13420, DBLP:conf/iclr/RenG0022, DBLP:conf/nips/Arjona-MedinaGW19}. These methods use a Markovian proxy reward function to replace environmental rewards, converting the delayed reward problem into a dense reward problem. Inspired by this progress, we consider addressing the challenge of sparse or delayed rewards in multi-agent tasks.

In this paper, we consider a new task called the spatial-temporal return decomposition problem, in which the algorithm must not only decompose the delayed episodic return to each time step but also assign credits to each agent according to their contribution. To solve this problem, we propose a novel approach called STAS, which establishes a connection between credit assignment and return decomposition. The Shapley value~\citep{shapley1997value,bilbao2000shapley,sundararajan2020many} is a well-known concept for computing relationships in cooperative games, as it considers all possible orderings in which players could join a coalition and determines the average marginal contribution of each player across all possible orderings. 
Inspired by that, STAS first uses a temporal attention to analyze state correlations within episodes, providing essential information for calculating Shapley values. Then it employs spatial attention module that uses Shapley values to allocate global rewards based on each agent's contribution.
We also introduce a random masked mechanism to overcome the computational complexity of the Shapley value by simulating different coalitions. In the spatial-temporal return decomposition problem, pinpointing key frames and the most contributory agent is vital for clear rewards that highlight actions positively impacting the final goal.
By using a dual transformer structure, STAS can identify key steps with the temporal transformer and use the spatial transformer to find the key agents. This enables STAS to obtain a Markovian reward, resulting in a better simulation of dense reward situations. 
Finally,  with successfully decomposed delayed return, single-agent can be effectively trained using policies such as  PPO~\cite{schulman2017proximal} or  SAC~\cite{haarnoja2018soft}.

In summary, our contributions are three-fold. 
Firstly, we define the Spatial-Temporal Return Decomposition Problem and propose STAS, which connects credit assignment with return decomposition to solve the problem.
Secondly, we introduce a dual transformer structure that utilizes a temporal transformer to identify key steps in long-time intervals and a spatial transformer to find the most contributory agents. We also leverage the Shapley value to compute the contribution of each agent and design a random masked attention mechanism to approximate it.
Lastly, we validate our approach in a challenging environment, Alice \& Bob, showing superior performance and effective cooperative pattern learning. Additionally, our method excels in general tasks in the Multi-agent Particle Environment (MPE), surpassing all baselines.

\section{Related Work}


\textbf{Return decomposition.}
Return decomposition, or temporal credit assignment, seeks to partition delayed returns into individual step contributions.
Earlier methods relied on naive principles, such as count-based \cite{ DBLP:conf/nips/Gangwani0020, DBLP:journals/corr/abs-1905-13420} or return-equivalent contribution analysis \cite{DBLP:conf/nips/Arjona-MedinaGW19, DBLP:conf/icml/PatilHDDBBAH22,zhang2023interpretable}. 
\citet{DBLP:conf/icml/HanRW0022} remodeled the problem and divided the value network into a history return predictor and a current reward one as HC-decompostion. \citet{DBLP:conf/iclr/RenG0022} proposed a concise method to utilize Monte Carlo Sampling as a time step sampler to decrease the computing complexity as well as the high variance. 
Although \citet{she2022agent} initially addressed sparse rewards in credit assignment, their method falls short in extremely delayed reward scenarios, with only terminal rewards at the trajectory's end.
More recently, \citet{DBLP:conf/atal/XiaoRP22} utilized attention structure to expand the return decomposition to multi-agent tasks as well as assign the reward to each agent in an implicit way by an end-to-end model.

\textbf{Credit assignment.}
Credit assignment methods can be divided into implicit and explicit methods. Implicit methods treat the global value and the local value as a whole for end-to-end learning. Earlier methods (\cite{DBLP:conf/atal/SunehagLGCZJLSL18, DBLP:conf/nips/RashidFPW20}) developed a network to aggregate the Q-values of each agent into a global Q-value based on additivity or monotonicity. Subsequent methods attempt to delve deeper into the characteristics of IGM to achieve more accurate estimation \cite{DBLP:conf/icml/SonKKHY19, DBLP:conf/iclr/WangH0DZ21} or employ other approaches such as Taylor expansion to estimate global value \cite{DBLP:journals/corr/abs-2002-03939, DBLP:conf/icml/0001DLZ20}. 

On the other hand, explicit methods attempt to establish an explicit connection between the global value and local value, enhancing the interpretability of the model by introducing algebraic relationships. \citet{DBLP:conf/aaai/FoersterFANW18} compute each agent's advantage with a counterfactual baseline, while \citet{DBLP:conf/aaai/WangZKG20} introduced Shapley value to calculate the contribution of each agent. \citet{DBLP:conf/kdd/LiKWLCWX21} combined counterfactual baseline and Shapley value to calculate each agent's contribution in different coalitions. \citet{du2019liir}  learns per-agent intrinsic reward functions to evade credit assignment issue.
Explicit methods typically employ the Actor-Critic architecture, which allows us to observe the contribution of each agent to the global reward, but this may result in high computational complexity.

\section{Preliminaries}
\textbf{Cooperative Multi-Agent Reinforcement Learning} 
\quad\quad In a fully cooperative multi-agent task ~\citep{panait2005cooperative}, agents need to work independently to achieve a common goal. Usually it can be described as a tuple: $\langle \mathcal{N}, \mathcal{S}, \mathcal{U}, P, r, \gamma, \rho_0\rangle$. Let $\mathcal{N} = \{1, 2, \cdots, n\}$ denote the set of $n$ agents. Denote the joint state space of the agents as $\mathcal{S} = \prod_{i=1}^n S_i$ and the joint action space of the agents as $\mathcal{U} = \prod_{i=1}^n U_i$ respectively. $P: \mathcal{S} \times \mathcal{U} \times \mathcal{S} \to [0, 1] $ is the state transition function. At time step $t$, let $\boldsymbol{s}_t = \{s^i
_t\}^n_{i=1}$ with each $s^i_t \in S_i$ being the state from agent $i$. Accordingly, let $\boldsymbol{u}_t = \{u^i
_t\}^n_{i=1}$ with each $u^i_t \in U_i$ indicating the action taken by the agent $i$. All agents will share a global reward from environment $r(\boldsymbol{s}_t, \boldsymbol{u}_t): \mathcal{S} \times \mathcal{U} \to \mathbb{R}$. $\gamma \in [0, 1)$ is a discount factor and $\rho_0 : \mathcal{S} \to \mathbb{R} $ is the distribution of the initial state $\boldsymbol{s}_0$. The goal of the agents is to find optimal policies that achieve maximum global \textit{return}, $J = \mathbb{E}_{\boldsymbol{s}_0,\boldsymbol{u}_0,\cdots} [\sum_{t=0}^T \gamma^t r_t]$ where $\boldsymbol{s}_0 \sim \rho_0(\boldsymbol{s}_0)$ and $T$ is the length of horizon. 

\textbf{Shapley Value}
Shapley value is a popular method for computing individual contributions in cooperative games~\citep{shapley1997value,bilbao2000shapley,sundararajan2020many}. Given a coalitional game $(v,N)$ where $N$ is the number of players cooperating together and $v$ refers to the value function describing the payoff that a coalition can obtain. For a particular player $i$, let $C$ denote an arbitrary subset that does not contain player $i$ and $C\cup \{i\}$ denote the attendance of player $i$ in subset $C$. Then the marginal contribution of player $i$ in $C$ can be defined as:
\begin{equation}
    v_i(C) = v(C\cup \{i\})-v(C).
    \label{eq:marginal_contri}
\end{equation}
The Shapley Value of player $i$ can be computed from the marginal contribution of player $i$ in all subsets of $N$:
\begin{equation}
    \Phi_v(i)=\underset{C\subseteq N\backslash \{i\}}{\sum} \frac{|C|!(|N|-|C|-1)!}{|N|!}v_i(C).
    \label{eq:shapley value}
\end{equation}
One crucial characteristic of the Shapley value is its efficiency property, which guarantees that the sum of the Shapley values of all agents equals the value of the grand coalition, i.e., $v(N)=\sum_{i=1}^N \Phi_v(i)$.

\textbf{Return Decomposition In Episodic Reinforcement Learning.} 
Commonly, agents receive a reward $r_t$ immediately after execution of action $\boldsymbol{u}_t$ at state $\boldsymbol{s}_t$. However, in the setting of episodic reinforcement learning~\cite{liu2019sequence, DBLP:conf/atal/XiaoRP22}, agents can only obtain one global reward feedback at the end of the trajectory. Let $\tau=(\boldsymbol{s}_0, \boldsymbol{u}_0, \boldsymbol{s}_1, \boldsymbol{u}_1,\cdots, \boldsymbol{s}_T)$ denote a trajectory of length $T$. We assume all trajectories terminate in finite steps. Then the episodic return function $r_{\text{ep}}(\tau)$ is defined on the trajectory space. The goal of episodic reinforcement learning in the multi-agent setting is to maximize the trajectory \textit{return}, $J = \mathbb{E}_{\tau=(\boldsymbol{s}_0,\boldsymbol{u}_0,\cdots)} [r_{\text{ep}}(\tau)]$. Therefore, the extremely sparse rewards will introduce large bias and variance~\citep{DBLP:conf/nips/Arjona-MedinaGW19,ng1999theory} into the process of training, let alone the lower sample efficiency when agents share such global rewards. 
In practice, we assume that the episodic return has some structure in nature, e.g., a sum-decomposable form~\citep{DBLP:conf/iclr/RenG0022}: 
\begin{equation}
    r_{\text{ep}}(\tau)\approx\hat{r}_{\text{ep}}(\tau)=\sum_{t=0}^{T-1}\hat{r}(\boldsymbol{s}_t,\boldsymbol{u}_t).
    \label{eq:return_decompose}
\end{equation}
This assumption is quite feasible and widely adopted in empirical studies~\citep{DBLP:journals/corr/abs-1905-13420}, where the task can be quantified by an additive metric, e.g., number of defeated enemies or team score in a sports match.

\section{Method}
\begin{figure*}[t]
    \centering
    \includegraphics[width=0.9\textwidth,height=0.55\textwidth]{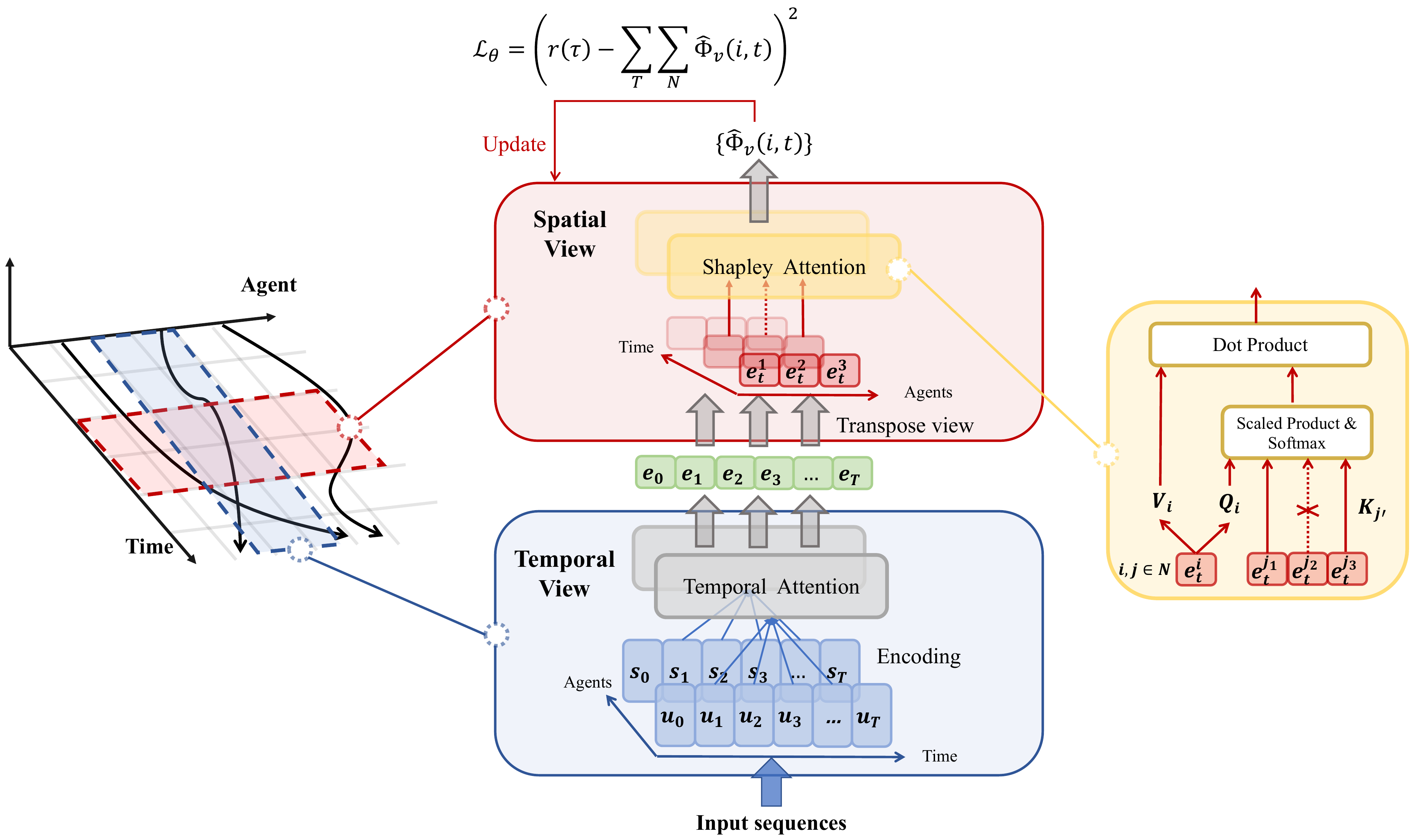}
    \caption{STAS framework. It contains a temporal attention module and a spatial Shapley attention module. Global states and actions of the entire episode are encoded and fed into the temporal attention module, which uses position embedding and time causality mask. The spatial Shapley attention module then approximates the Shapley value from previously learned representations in the spatial dimension. Finally, the model is updated based on the Shapley value approximation.}
    \label{fig:framework}
\end{figure*}

In this paper, we address the challenge of fully cooperative multi-agent systems with episodic global return. We propose the STAS framework in Fig~\ref{fig:framework}, which decomposes the return in both temporal and spatial dimensions.

Following Eq. \eqref{eq:return_decompose}, we assume that the sum of dense global rewards along the episode can reconstruct the final return given at the last time step. Similarly, the global reward equates to the sum of all distributed individual rewards.
\begin{equation}
    r_{\text{ep}}(\tau)=\sum_{t=0}^{T-1}r(\boldsymbol{s}_t,\boldsymbol{u}_t)\\
    =\sum_{t=0}^{T-1}\sum_{i=1}^N \Phi_v(i,t).
    \label{eq:episodic_return}
\end{equation}
where $\Phi_v(i,t)$ is the distributed reward for each agent. As mentioned previously, the use of Shapley Value in cooperative games has proven beneficial, distributing the payoff of each agent and leading to faster convergence rates. Consequently, we calculate the Shapley value as the actual distributed reward for each agent.
\subsection{Spatial Decomposition: Shapley Value}
As indicated in \eqref{eq:shapley value}, the Shapley Value of agent $i$ at time $t$ is computed by iterating through all possible subsets of the grand coalition:
\begin{equation}
\Phi_v(i,t)=\underset{C\subseteq N\backslash {i}}{\sum} \frac{|C|!(|N|-|C|-1)!}{|N|!}v_i(C,t),
\end{equation}
where $N\backslash {i}$ represents subsets that exclude agent $i$, and $v_i(C,t)$ is the marginal contribution of agent $i$ for a coalition $C$ at time $t$:
\begin{equation}
    v_i(C,t) = v(\boldsymbol{s}_{C\cup \{i\},t},\boldsymbol{u}_{C\cup \{i\},t})-v(\boldsymbol{s}_{C,t},\boldsymbol{u}_{C,t}),
\end{equation}
where $\boldsymbol{s}_{C'}=(s_i)_{i\in C'}$ and $\boldsymbol{u}_{C'}=(u_i)_{i\in C'}$. 

\textbf{Marginal Contribution Approximation} \quad In practice, it is difficult and unstable to either learn a global value function for each time step $t$ or a local value function for arbitrary coalition. When the number of agents increases, the computational complexity grows exponentially. Therefore, we propose a simple method to approximate the marginal contribution. Suppose all agents take actions sequentially to join a grand coalition $C$. Then we define a function to estimate marginal contribution as:
\begin{align}
    \hat{v}_i(C,t;\theta):\mathcal{S}_{C\cup \{i\}}\times \mathcal{U}_{C\cup \{i\}}\to \mathbb{R}, \nonumber
\end{align}
where $\mathcal{S}_{C\cup \{i\}}=(\mathcal{S}_j)_{j\in C\cup \{i\}}$, $\mathcal{U}_{C\cup \{i\}}=(\mathcal{U}_j)_{j\in C\cup \{i\}}$, $C$ is an ordered coalition that agent $i$ is about to join and $\theta$ is the parameters of the function. It is easy to extend the above definition to the game where agents take actions simultaneously, by taking expectation over all possible permutations of coalition $C$. In practice, we use self-attention~\citep{shaw2018self} to implement such marginal contribution approximation. The query vector for agent $i$ at time step $t$ is denoted as $\boldsymbol{q}^{\text{SPA}}_{i,t}=\boldsymbol{W}_Q^{\text{SPA}}\boldsymbol{e}_{i,t}^\mathrm{T}$, where $\boldsymbol{e}_{i,t}$ represents the encoding of the state and action of agent $i$ at time step $t$, and $\boldsymbol{W}_Q^{\text{SPA}}$ is a trainable query projection matrix. Similarly, the key and value vectors for agent $j$ at time $t$ are denoted as $\boldsymbol{k}^{\text{SPA}}_{i,t}=\boldsymbol{W}_K^{\text{SPA}}\boldsymbol{e}_{i,t}^\mathrm{T}$ and $\boldsymbol{v}^{\text{SPA}}_{j,t}=\boldsymbol{W}_V^{\text{SPA}}\boldsymbol{e}_{j,t}^\mathrm{T}$, respectively. The self-attention weight can be computed as follows:
\begin{equation}
    \alpha_t^{ij}=softmax(\frac{\boldsymbol{q}^{\text{SPA}}_{i,t}{\boldsymbol{k}^{\text{SPA}}_{j,t}}^\mathrm{T}}{\sqrt{d}}),
\end{equation}
where $d$ represents the dimension of the $\boldsymbol{Q}$ and $\boldsymbol{K}$. The attention weight $\alpha_t^{ij}$ can be interpreted as a correlation between agent $i$ and $j$ at time $t$. When viewed from the perspective of agent $i$, $\alpha_t^{ij}=0$ signifies that agent $j$ is not in the coalition that agent $i$ is preparing to join. Thus, a straightforward mask $\boldsymbol{M}_i\in\mathcal{R}^N$ on other agents 
 is sufficient to model the process of agent $i$ joining arbitrary coalition $C$:
\begin{equation}
    \boldsymbol{M}_{i}[j]= \left\{
    \begin{aligned}
    & 1 , j\in C\cup \{i\} \\
    & 0, j\notin C\cup \{i\}.
    \end{aligned}
\right.\\
\label{eq:agents_mask}
\end{equation}
 After masking irrelevant agents, the masked attention function in a matrix form:
 \begin{equation}
 \label{eq:maskedattention}
     \operatorname{MA}(\boldsymbol{q}^{\text{SPA}}_{i,t},\boldsymbol{K}^{\text{SPA}}_{t},\boldsymbol{V}^{\text{SPA}}_{t})=
     \operatorname{softmax}(\frac{\boldsymbol{q}^{\text{SPA}}_{i,t}{\boldsymbol{K}^{\text{SPA}}_{t}}^\mathrm{T}}{\sqrt{d}}\odot \boldsymbol{M}_i)\boldsymbol{V}^{\text{SPA}}_{t},
 \end{equation}
 where $\odot$ is a Hadamard product. And it is reasonable to represent the approximation of the marginal contribution of agent $i$, i.e., $\hat{v}_i(C,t; \theta)=\operatorname{MA}(\boldsymbol{q}^{\text{SPA}}_{i,t},\boldsymbol{K}^{\text{SPA}}_{t},\boldsymbol{V}^{\text{SPA}}_{t})$. We believe that such modeling can maintain the property of exact marginal contribution.
 
\textbf{Shapley Value Approximation}
 \quad As the number of agents in a multi-agent system grows, computing the exact Shapley Value for each agent becomes increasingly difficult, even when using marginal contribution approximation. To address this challenge, recent studies ~\citep{ghorbani2019data,maleki2013bounding} have proposed approximate methods as an alternative to the exact computation of Shapley Value. These approximation methods allow for more efficient estimation of the Shapley Value, while still capturing the essential information about each agent's contribution to the team's performance. To mitigate the unacceptable computational burden brought by traversing all possible coalitions, we also adopt Monte Carlo estimation for Shapley Value approximation:
 \begin{equation}
     \hat{\Phi}_v(i,t;\theta)=\frac{1}{K}\sum_{k=1}^K \hat{v}_i(C_k,t;\theta),\forall C_k \subseteq N\backslash \{i\},
 \end{equation}
 Although other estimation methods are available \cite{zhou2023probably}, for the sake of simplicity and the absence of prior knowledge, we deem Monte Carlo estimation to be both necessary and sufficient for spatial reward decomposition.
\subsection{Temporal Decomposition}
Due to the efficiency property of the Shapley Value, the sum of all distributed payoffs equals the value of grand coalition. Therefore, we can get the estimated global reward at time $t$ from the Shapley Value approximation of each agent, and the estimated final return can be written as:
\begin{equation}
    \hat{r}_{\text{ep}}(\tau)=\sum_{t=0}^{T-1}\sum_{i=1}^N \hat{\Phi}_v(i,t;\theta).
    \label{eq:estimated_return}
\end{equation}
In fact, the marginal contribution of an agent is seldom independent across time unless the relevant information from previous states has already been considered. To decompose the episodic return into a Markovian proxy reward function, the relations between actions and state transitions along trajectories must be established. 

In practice, sequential modeling methods like Long short-term memory (LSTM) ~\citep{sak2014long} or Transformer ~\citep{vaswani2017attention} are favored to encode the trajectory. Similar to Eq.~\eqref{eq:maskedattention}, we adopt an attention function with causality masks on the state-action encoding $\boldsymbol{e}$ to incorporate temporal information. Specifically, we have
\begin{equation}
\begin{split}
    \boldsymbol{e}^i_t &= \operatorname{MA}(\boldsymbol{q}^{\text{TEM}}_{i,t},\boldsymbol{K}^{\text{TEM}}_{i},\boldsymbol{V}^{\text{TEM}}_{i})\\
    &=\operatorname{softmax}(\frac{\boldsymbol{q}^{\text{TEM}}_{i,t}{\boldsymbol{K}^{\text{TEM}}_{i}}^\mathrm{T}}{\sqrt{d}}\odot \boldsymbol{M}^{\text{TEM}})\boldsymbol{V}^{\text{TEM}}_{i},\\
\end{split}
\end{equation}
where the query matrix $\boldsymbol{q}^{\text{TEM}}_{i,t}$ is obtained by applying the trainable matrix $\boldsymbol{W}_Q^{\text{TEM}}$ to the output of a feed-forward layer $f$ that encodes raw state and action embeddings. Similarly, the key and value matrices are generated by the trainable matrices $\boldsymbol{W}^{\text{TEM}}_K$ and $\boldsymbol{W}^{\text{TEM}}_V$, respectively. The self-attention weight is calculated using a softmax function applied to the scaled dot product of the query and key matrices, with the scaling factor $\sqrt{d}$ and the causality mask $\boldsymbol{M}^{\text{TEM}}$. $\boldsymbol{M}^{\text{TEM}}$ is a causality mask with its first $t$-th entries equal to $1$, and remaining entries $0$. It ensures that at each time step $t$, the model only attends to the states and actions before $t$.

\subsection{Overall objective}
With the estimated return from Eq.~\eqref{eq:estimated_return}, we can train the return decomposition model by the following objective:
\begin{equation}
    \mathcal{L}(\theta)=\underset{\tau \sim \mathcal{D}}{\mathbb{E}}\left[\biggl(r_{\text{ep}}(\tau) - \sum_{t=0}^{T-1} \sum_{i=1}^N \hat{\Phi}_v(i,t;\theta))\biggl)^{2}\right],
    \label{eq:loss}
\end{equation}
where $\hat{\Phi}_v(i,t;\theta)$ is the approximation of the Shapley Value of agent $i$ at time $t$, $\mathcal{D}$ denotes the trajectories collected in the experience buffer and $\theta$ denotes the parameter of the return decomposition model. The details of the training algorithm are shown in algorithm~\ref{tab:training}.

\textbf{Policy optimization}\quad For every policy, the complete trajectory's global states and actions can be fed into the return decomposition model to derive rewards for each agent at each time step, denoted as $\hat{\Phi}_v(i,t;\theta)$. Subsequently, the policy can be updated using algorithms like PPO~\cite{schulman2017proximal} or  SAC~\cite{haarnoja2018soft}.

\begin{algorithm}[htb]
    \caption{Spatial-Temporal Attention with Shapley value for Return Decomposition}
    \label{tab:training}
    \begin{algorithmic}[1]
    \REQUIRE number of agents $N$; Initialized parameters $\theta$ of return decomposition model; Initialized parameters $\phi_1,\cdots,\phi_N$ of policy networks respectively; Experience buffer $B\gets \emptyset$;  Return decomposition model training frequency $M$
    \FOR{$k=0,1,\cdots,K$}
    \STATE sample initial state $\boldsymbol{s}_0\sim\rho_0$
    \FOR{$t=0,1,\cdots,T-1$}
    \STATE sample actions $u_t^i\sim\pi_{\phi_i}(s_t^i)$, for $i=0,1,\cdots,N$
    \STATE Take joint actions $\boldsymbol{u}_t$ and observe next state $\boldsymbol{s}_{t+1}$
    \ENDFOR
    \STATE Collect episodic return $r_{\text{ep}}(\tau)$ and the trajectory $\tau=\{\boldsymbol{s}_0,\boldsymbol{u}_0,\cdots,\boldsymbol{s}_{T-1},\boldsymbol{u}_{T-1}\}$, store $(\tau, r_{\text{ep}})$ into the buffer $B$
    \STATE Sample a batch of trajectories from buffer $D\sim B$
    \STATE Predict the decomposition reward $\{\hat{\Phi}_v(i,t; \theta)\}$ for each $\tau\in D$ \STATE Update parameter $\phi_i$ with independent trajectories $\tau_i=\{(s_t^i,u_t^i,\hat{\Phi}_v(i,t; \theta))\}$, for $i=0,1,\cdots,N$
    \IF{$k$ mod $M$ is $0$}
    
    \REPEAT
    \STATE Sample a batch of trajectories from buffer $D'\sim B$ and compute loss with Eq.~\eqref{eq:loss}
    \STATE Update $\theta\gets \theta-\nabla_\theta\mathcal{L}(\theta)$ 
    \UNTIL{Convergence}
    \ENDIF
    \ENDFOR
    \end{algorithmic}
\end{algorithm}

\section{Experiments}
In this section, we conduct experiments to evaluate the performance and effectiveness of our proposed method on the challenging problem of spatial-temporal return decomposition. We evaluate our approach using two different implementations and test it on three different environments: a newly designed environment ``Alice \& Bob'', as well as the \textit{cooperative navigation} and \textit{predator-prey} scenarios in the general multi-agent benchmark MPE~\citep{lowe2017multi}. The source code is available at \texttt{https://github.com/zowiezhang/STAS}.
\subsection{Evaluate Algorithms}
We compared our method with other several baseline algorithms in a scenario where rewards are given only at the game's end. Agents receive a $0$ reward at each step and the sum of all rewards, i.e., the episodic return, after the game ends.
We have evaluated two implementations of our method:
\begin{itemize}
\item \textbf{STAS (ours)} denotes the default implementation of our algorithm. First, we input the state-action sequence into a temporal attention module, which is designed to extract essential information from previous states. Next, we use a Shapley attention module to approximate the Shapley value for each agent. The predicted Shapley value is a measure of an agent's instant contribution to the team's final performance.
\item \textbf{STAS}-\textbf{ML} \textbf{(ours)} is our alternative implementation, where ``ML'' stands for multi-layers. We organize temporal and spatial components into layers, each comprising a single temporal attention module and a single Shapley attention module. By stacking multiple layers, we construct our final prediction of the Shapley value. Each layer captures increasingly complex dependencies and interactions among agents, resulting in more accurate estimates of their individual contributions.
\end{itemize}
We compare with several existing baselines under the same settings, including \textbf{QMIX}~\citep{DBLP:conf/nips/RashidFPW20}, \textbf{COMA}~\citep{DBLP:conf/aaai/FoersterFANW18} and \textbf{SQDDPG} ~\citep{DBLP:conf/aaai/WangZKG20}
    

\textbf{The specific implementation of our methods} \quad 
We utilize a temporal attention module to extract crucial information from previous states. To ensure causality, each state only receives information from earlier states through a causality mask. Positional embedding is employed to capture the relative positions of each state in the sequence.
The Shapley attention module processes the learned embeddings for each agent to estimate their Shapley value. 
Our policy model is based on independent PPO ~\citep{schulman2017proximal}.
\begin{figure}[htb!]
\centering 

\includegraphics[width=0.5\linewidth,height=0.5\linewidth]{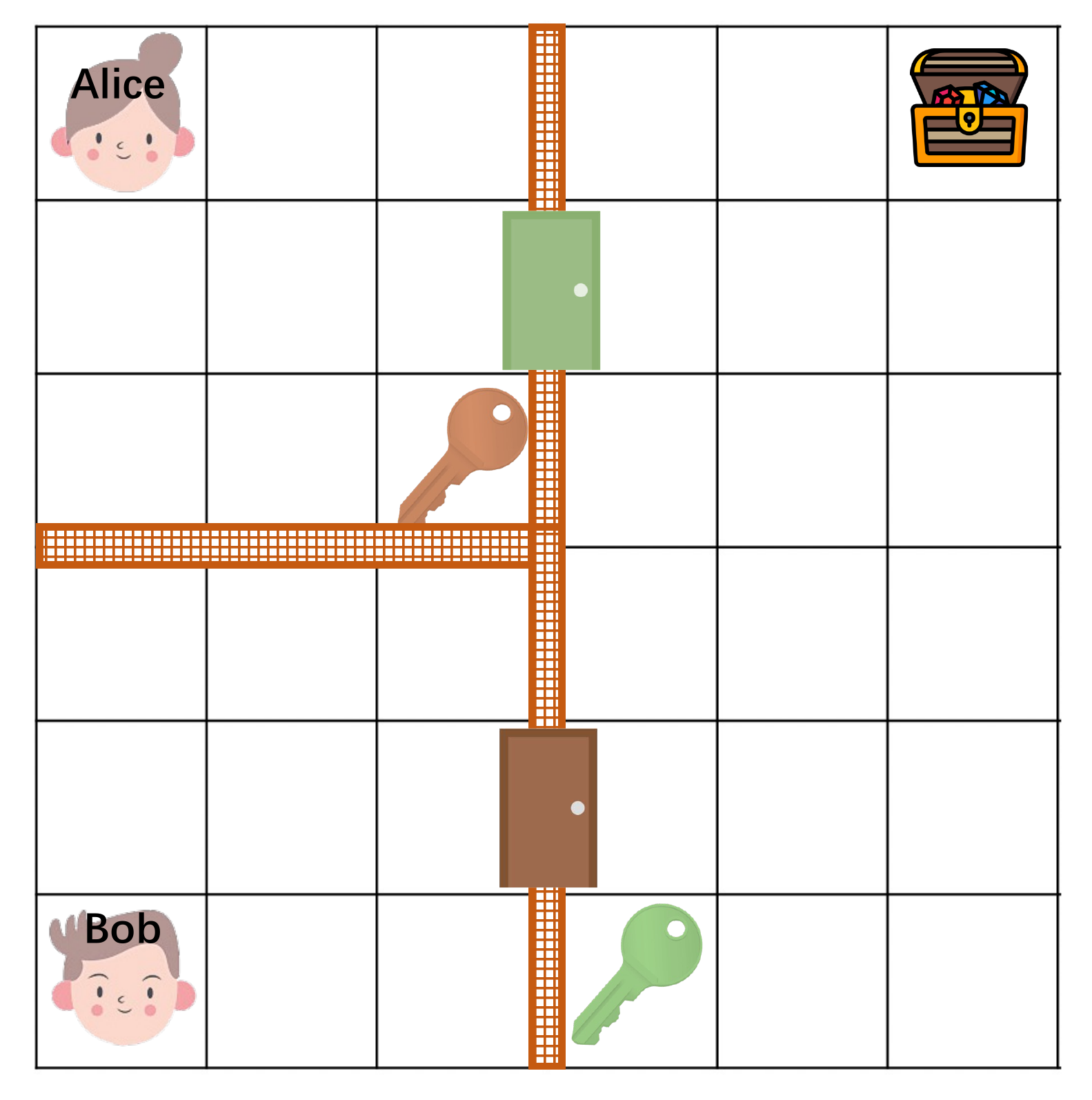}
\caption{A simple demonstration of the designed extreme delayed reward environment Alice \& Bob. When a key is reached, the corresponding door will open. To obtain the treasure, Alice must first retrieve the brown key to unlock the door to Bob's room. After Alice has unlocked the door, Bob can then retrieve the green key to open the door to Alice's room. With both doors unlocked, they can proceed to the treasure together.}
\label{A_B}
\end{figure}
\subsection{Extreme delayed reward environment}

In this experiment section, we designed a toy example environment called ``Alice \& Bob'' to verify the effectiveness of our method in an extremely delayed reward setting, as illustrated in Fig~\ref{A_B}. 
In this task, agents Alice and Bob, unable to communicate, must collaboratively secure a treasure within a set number of steps, following a specific order and relying solely on their individual environmental states. They must adapt to the cooperative pattern dictated by the environment. The task's complexity is heightened by an extremely delayed reward system, where rewards accumulate and are only dispensed upon task completion, posing a substantial challenge to solving methods.

\subsubsection{Environment settings}
At the beginning of the task, both Alice and Bob are locked in a room, but Alice can use the key in her room to unlock Bob's room. Once the key is touched, the door of Bob's room will be opened. Then Bob can get out of the room and find the key to Alice's room. After both of them get out of the room, they can go and get the treasure. The reward setting is that if any of them hit the wall when exploring, they will get a minus reward -0.2; if at the end of the game, they reach the treasure, they will get an extra reward of 200 and finish the task, and if they still not get the treasure within the maximum step, they will get the reward sum of each step as the whole return to force stop the task. 
 Each agent has 4 different actions, namely, move\_up, move\_down, move\_left and move\_right. 

\subsubsection{Results}
\begin{figure}[htb]
\centering

\subfigure{
\includegraphics[width=0.9\linewidth]{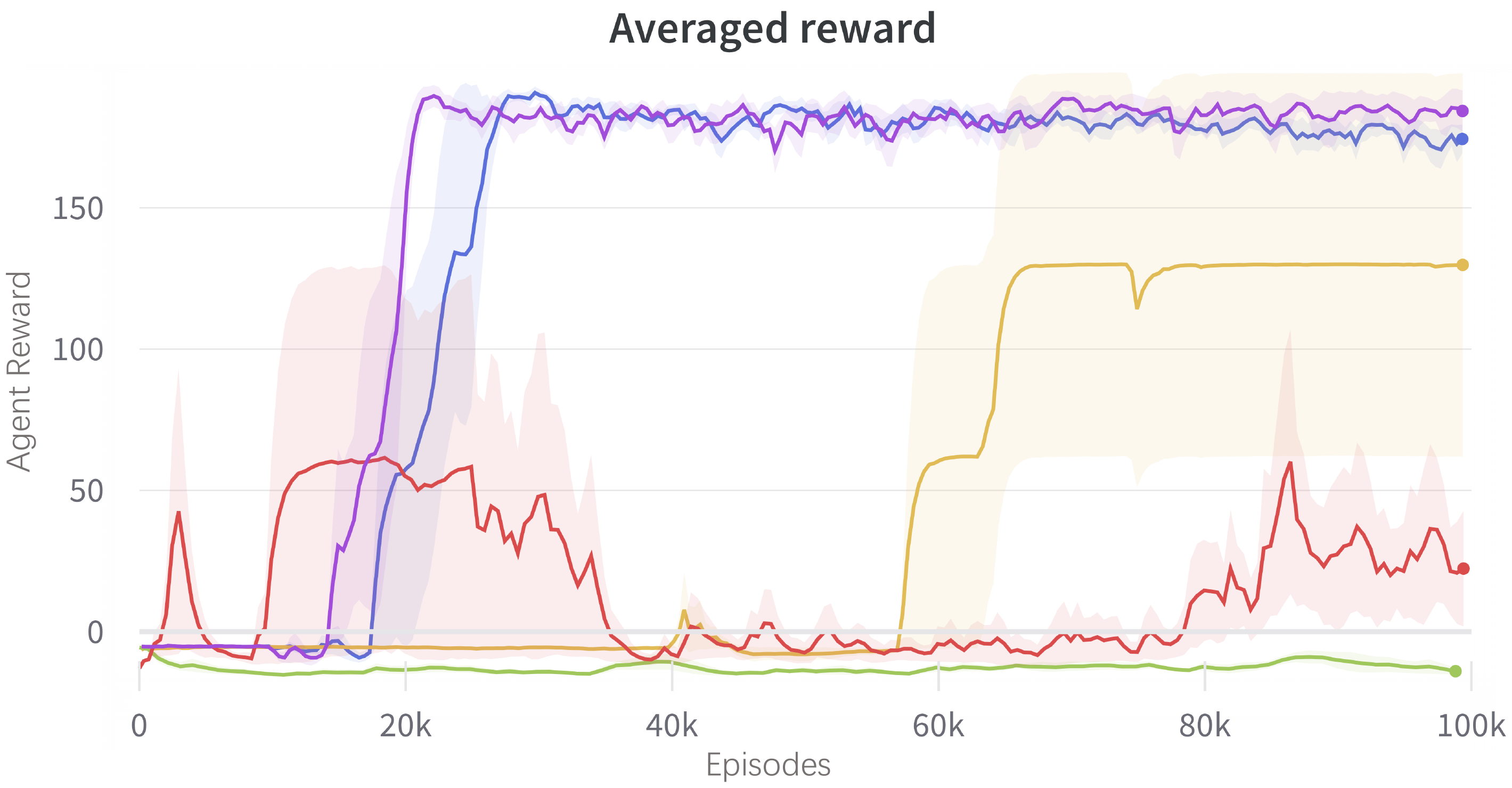} \label{Fig.ab(a)}
}

\subfigure{
\includegraphics[width=0.9\linewidth]{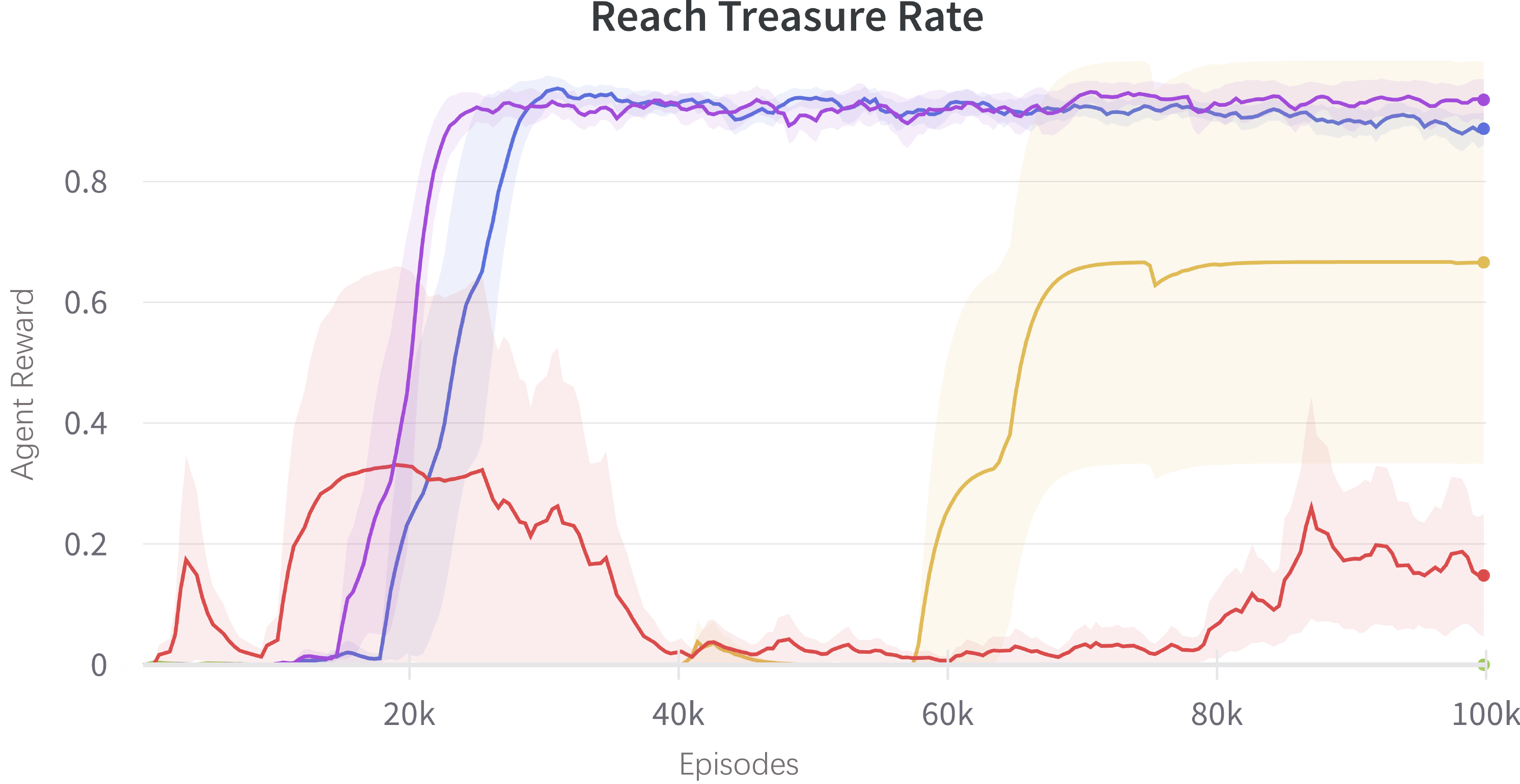} \label{Fig.ab(b)}
}	

\subfigure{
\includegraphics[width=0.8\linewidth]{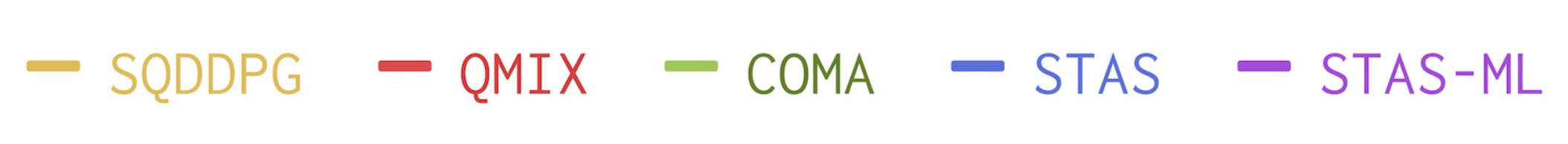}
}
\caption{Average agent rewards and reaching treasure rate with standard deviation for task Alice \& Bob.}

\label{Fig.ab}
\end{figure}
\begin{figure*}[h!]
\centering

\subfigure[Cooperative Navigation(3 agents)]{
\includegraphics[width=0.30\linewidth]{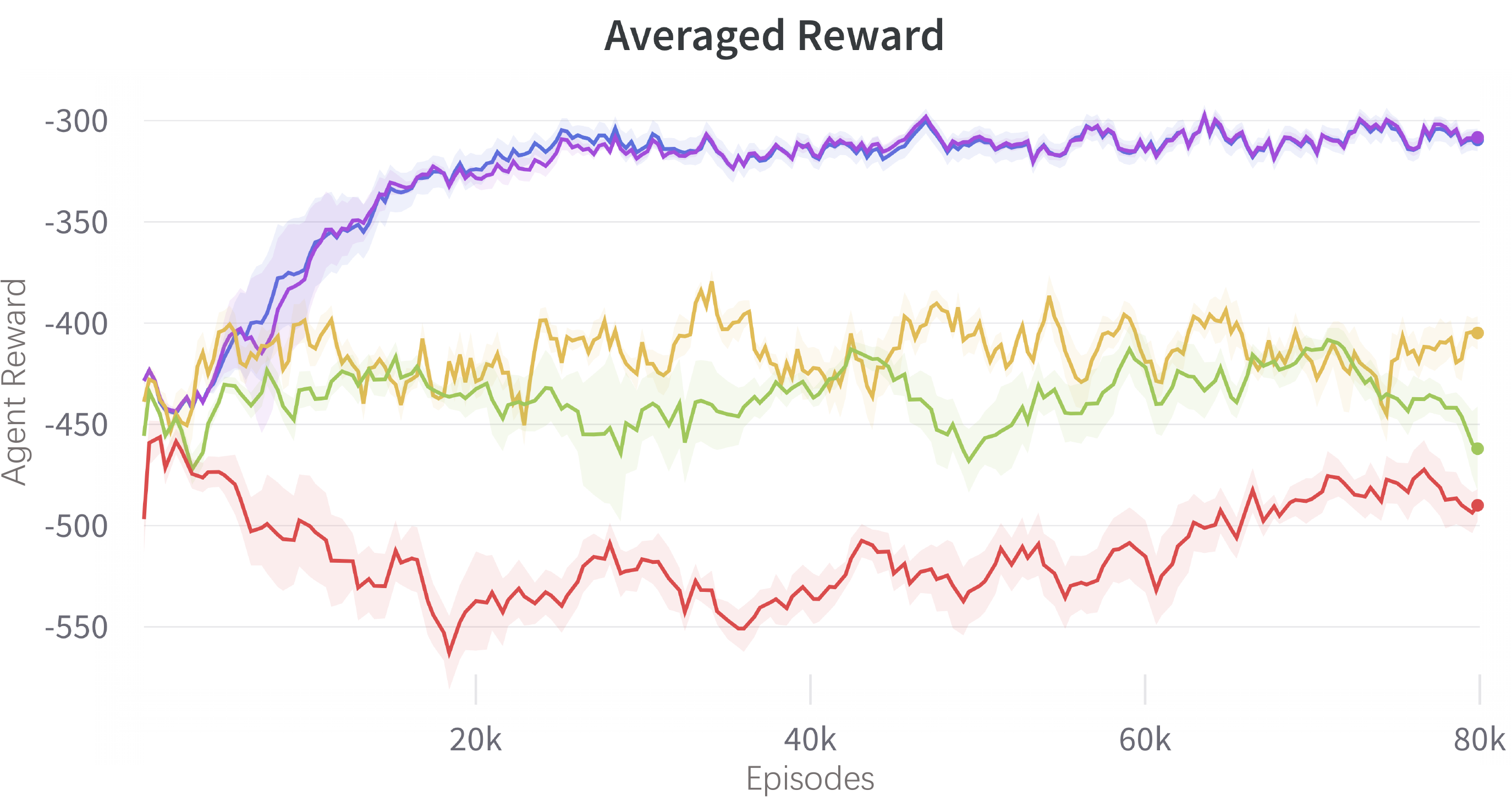} 
\label{fig:mpe_simple_spread_n3}
}
\subfigure[Cooperative Navigation(6 agents)]{
\includegraphics[width=0.30\linewidth]{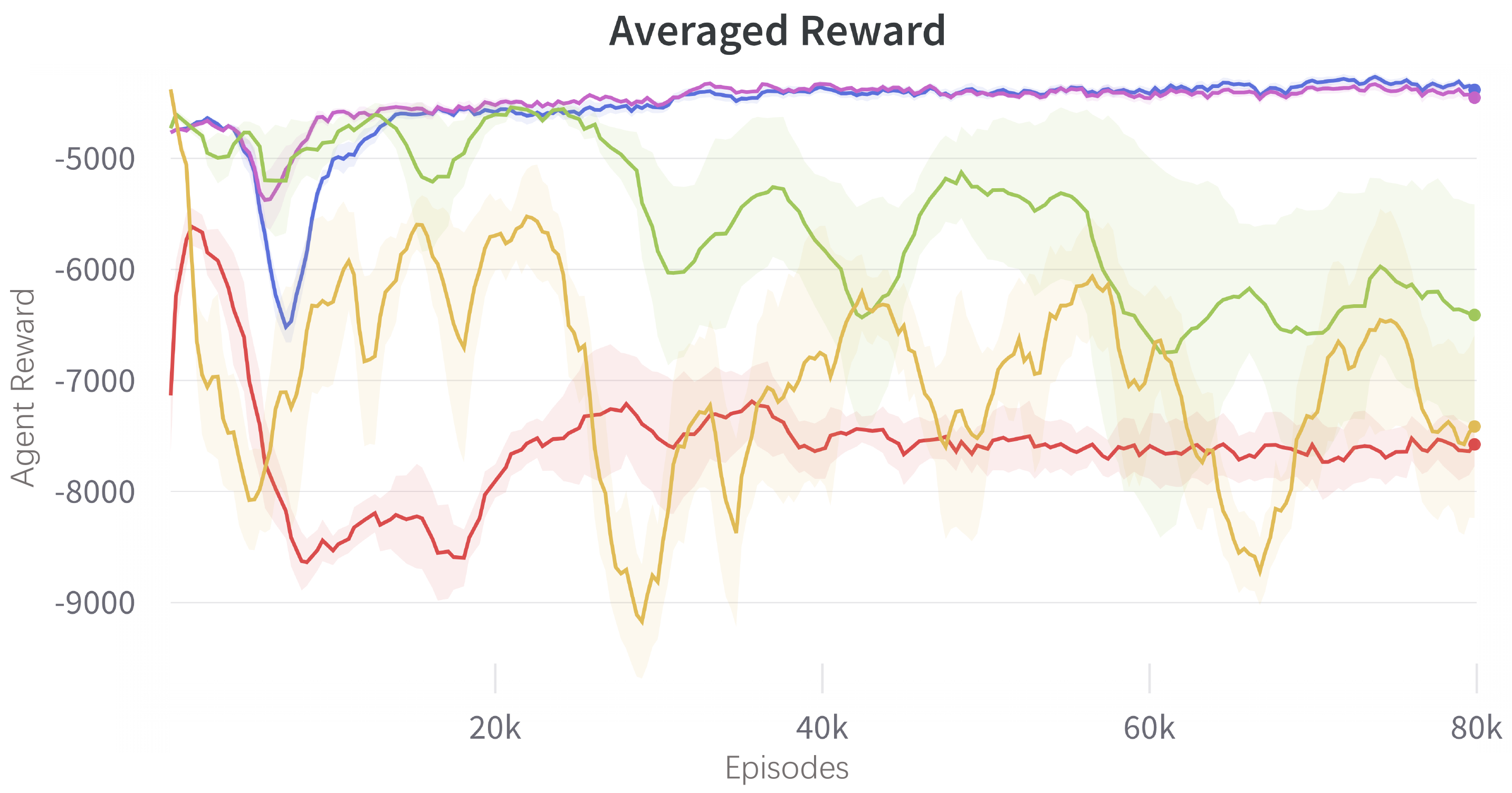} 
\label{fig:mpe_simple_spread_n6}
}
\subfigure[Cooperative Navigation(15 agents)]{
\includegraphics[width=0.30\linewidth]{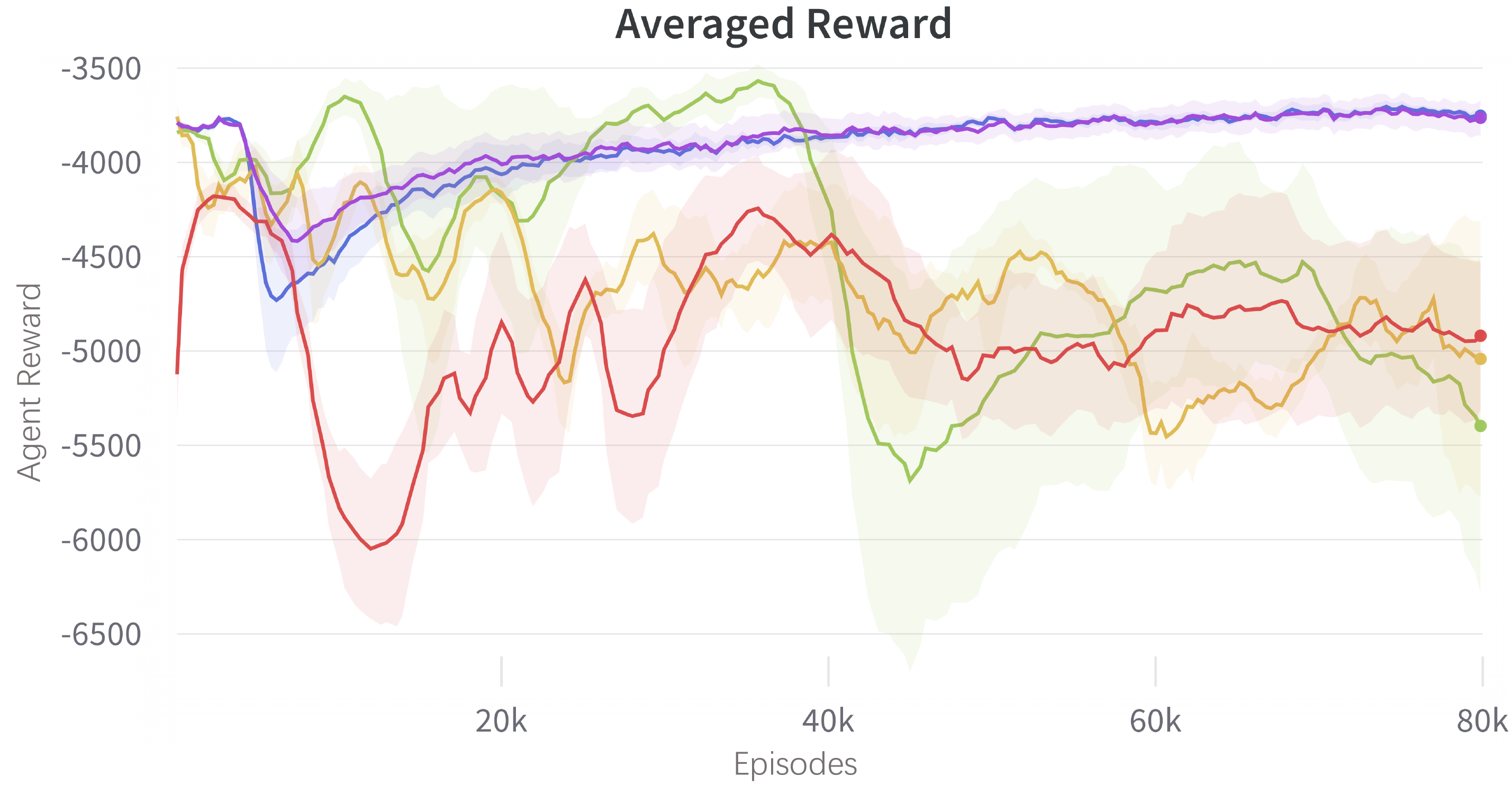} 
\label{fig:mpe_simple_spread_n15}
}

\subfigure[Predator-prey(3 agents)]{
\includegraphics[width=0.30\linewidth]{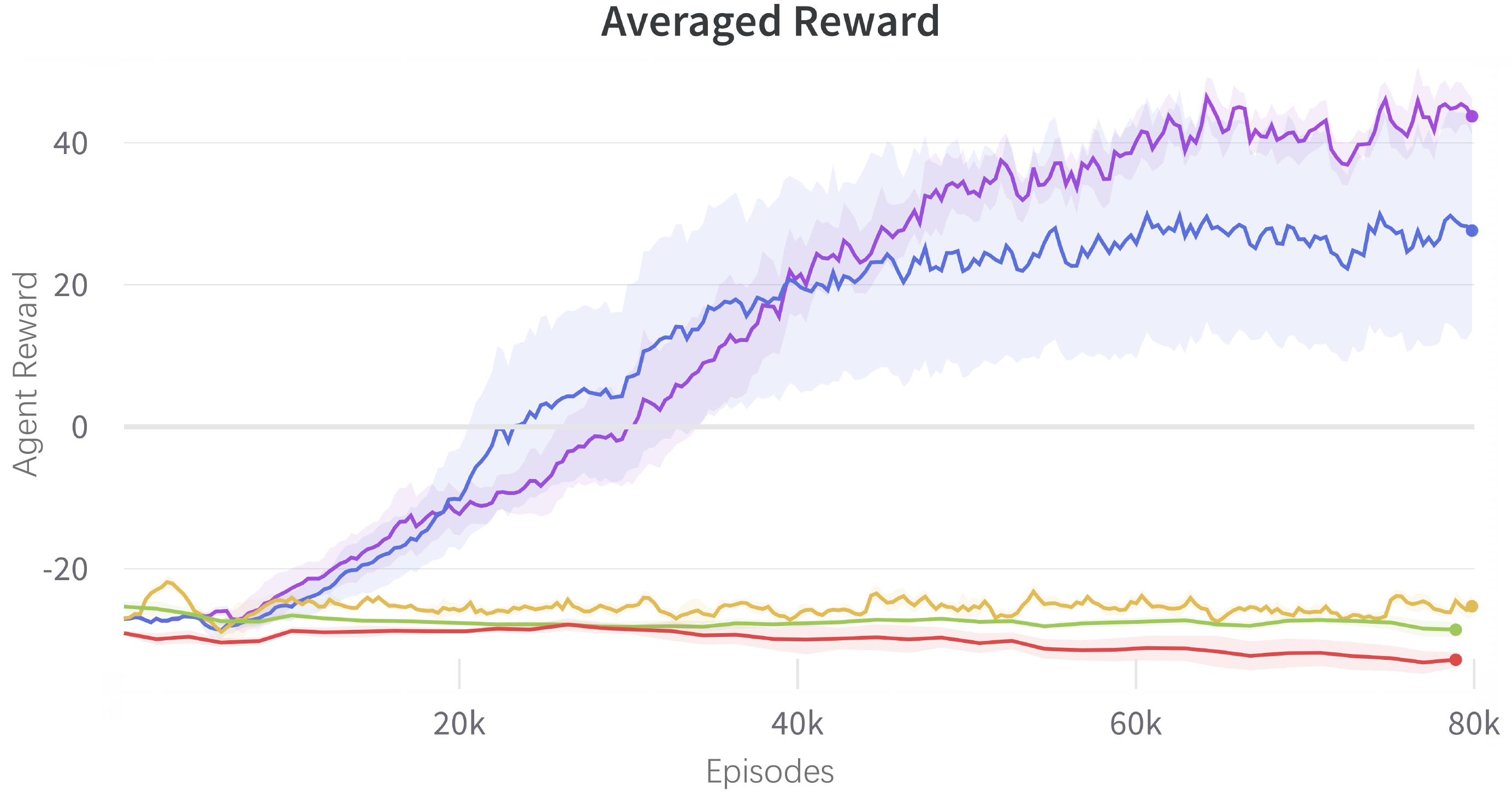} 
\label{fig:mpe_simple_tag_n3}
}
\subfigure[Predator-prey(6 agents)]{
\includegraphics[width=0.30\linewidth]{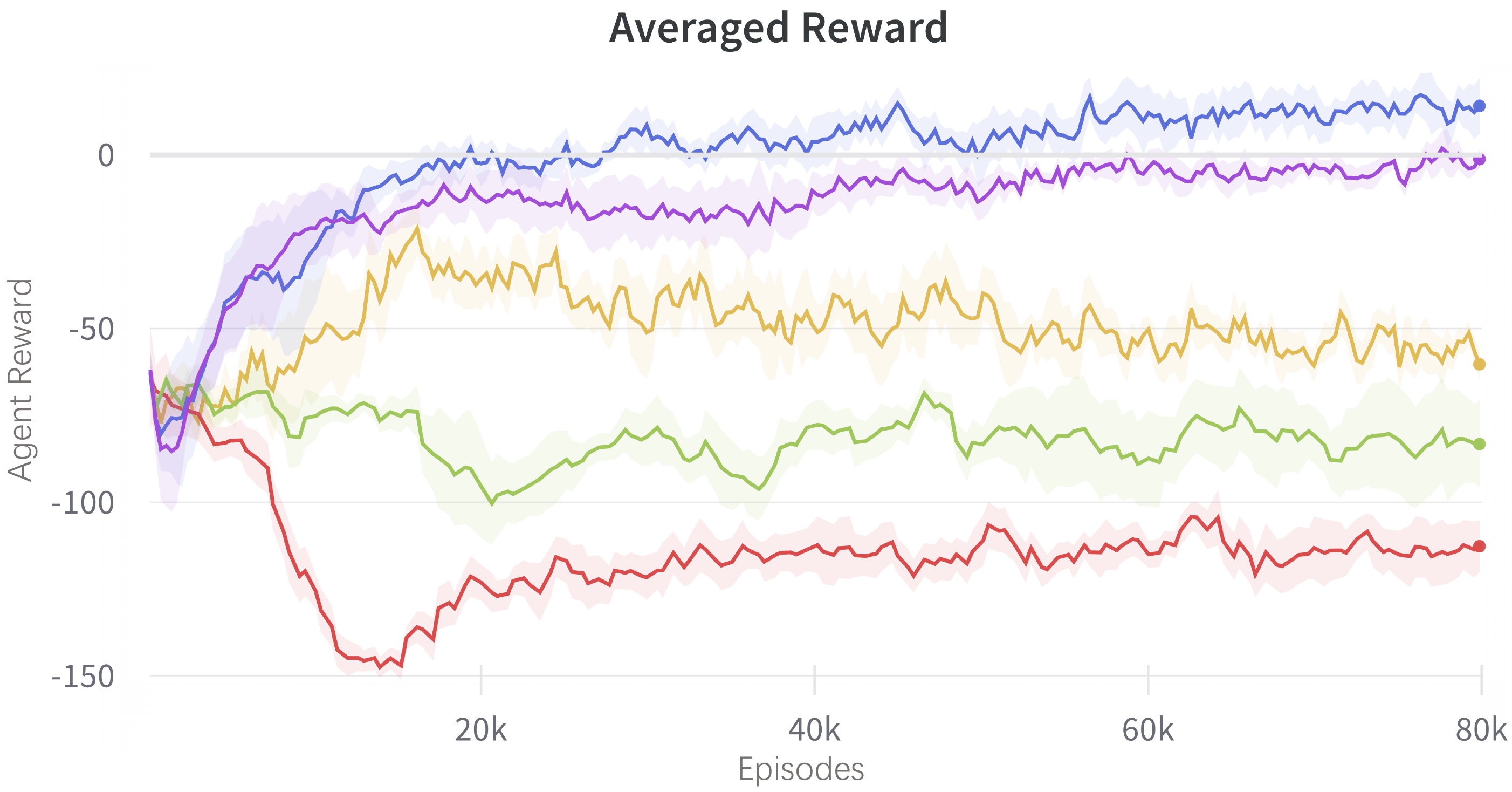} 
\label{fig:mpe_simple_tag_n6}
}
\subfigure[Predator-prey(15 agents)]{
\includegraphics[width=0.30\linewidth]{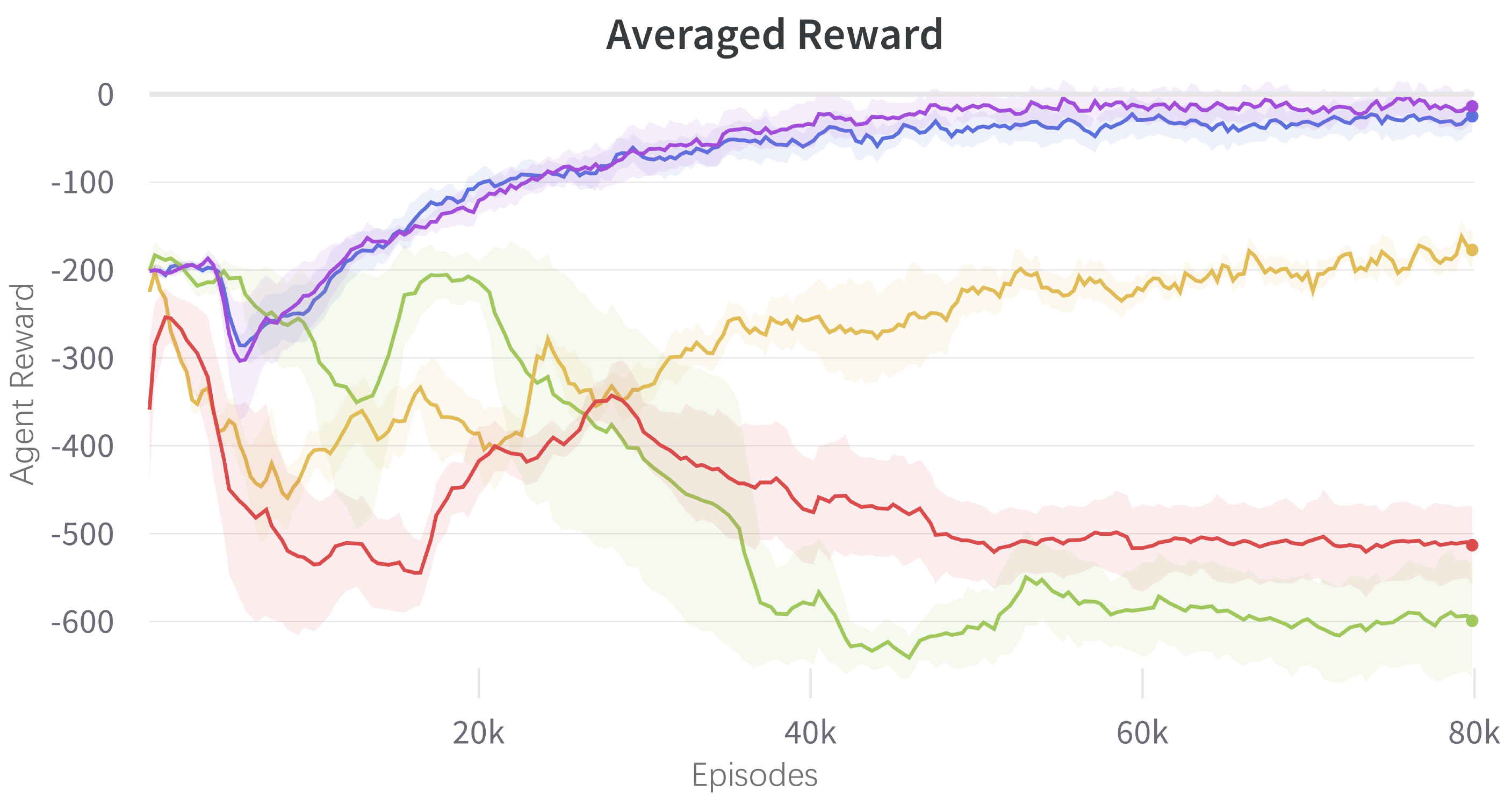} 
\label{fig:mpe_simple_tag_n15}
}

\subfigure{
\includegraphics[width=0.52\linewidth]{images/legend.pdf}
}
\caption{Average agent rewards with standard deviation for scenario Cooperative Navigation and Predator-prey in Multi-agent Particle Environments.}

\label{Fig:MPE}
\end {figure*}
\begin{figure*}
    \centering
    
    \subfigure[Different sample number]{
    \includegraphics[width=0.41\linewidth]{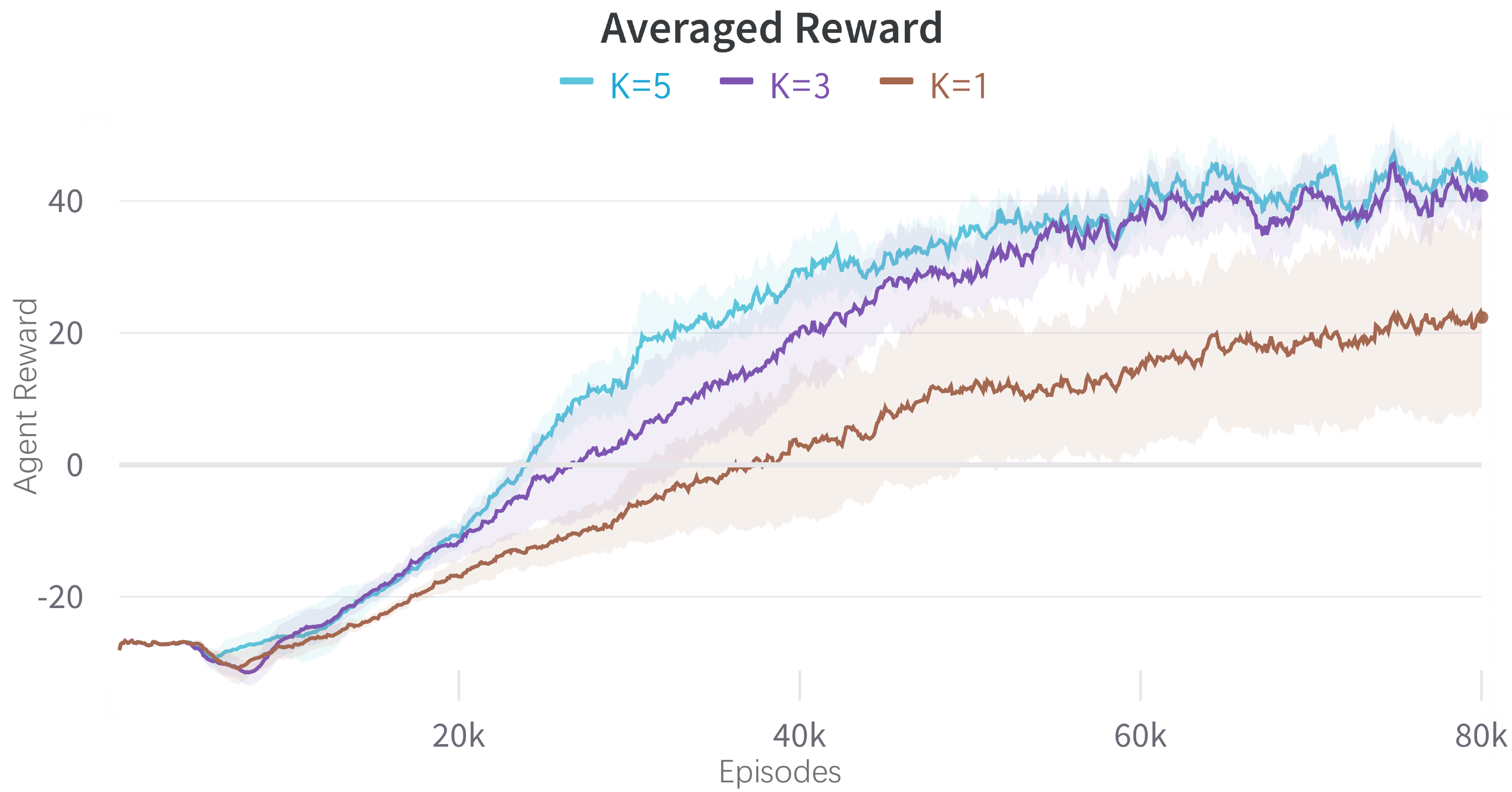}
    \label{fig:sample_num}
    }
    \subfigure[Different Layer number]{
    \includegraphics[width=0.41\linewidth]{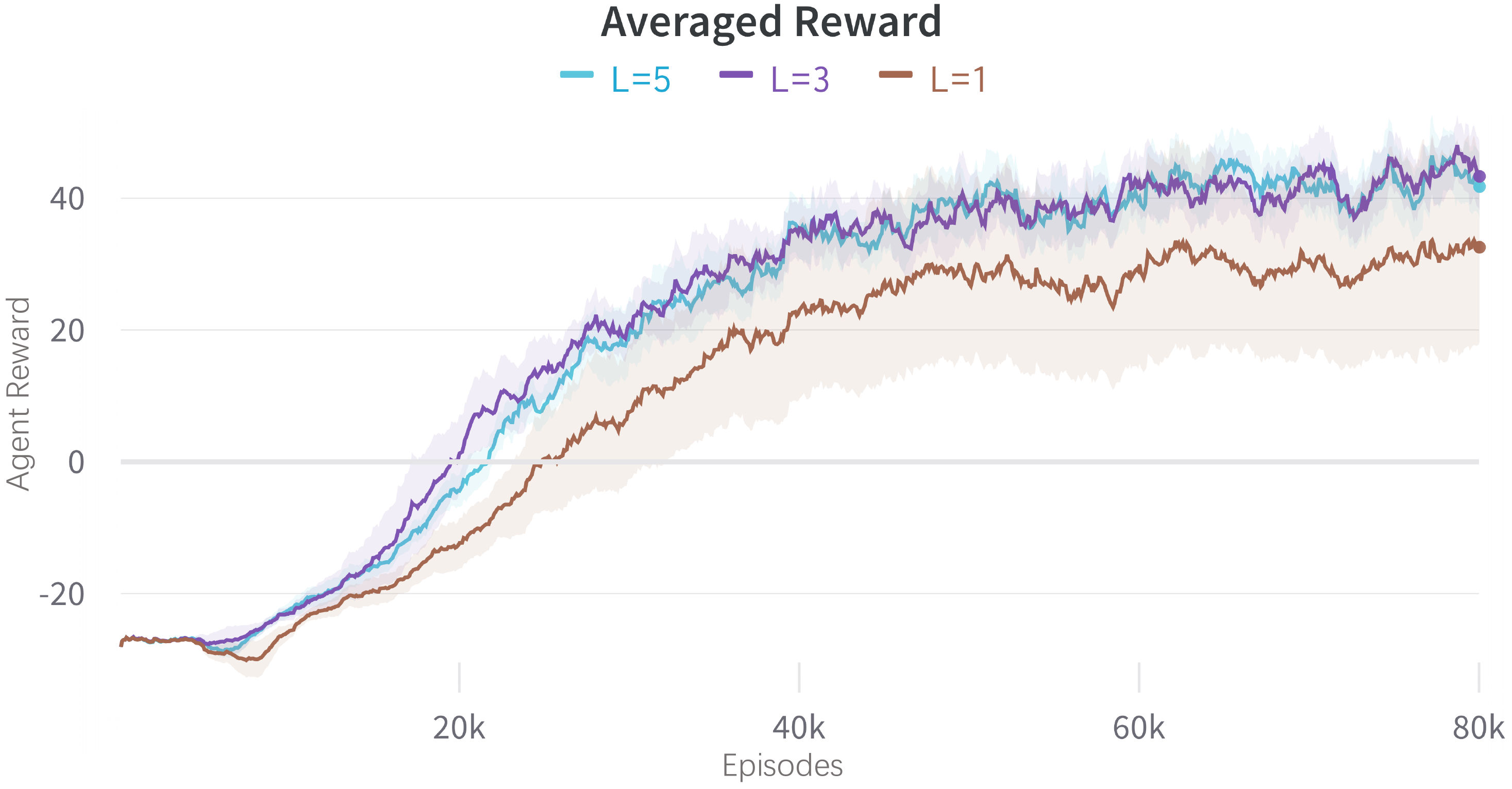}
    
    \label{fig:layer_num}
    }
    \caption{The average reward of STAS(a) and STAS-ML(b) in task \textit{predator-prey}(3 agents)}

\end{figure*}
\begin{table*}[htbp!]
\centering
\begin{tabular}{@{}lcccc@{}}
\toprule
& COMA & QMIX & SQDDPG & STAS \\
\midrule
Coefficient & \(3.5829 \times 10^{-4}\) & \(-0.1572\) & \(0.2672\) & \(0.3592\) \\
Two-tailed \(p\)-value & 0.9218 & \(5.1895 \times 10^{-43}\) & \(1.6786 \times 10^{-147}\) & \(4.2386 \times 10^{-210}\) \\
\bottomrule
\end{tabular}
\caption{The Pearson correlation coefficient between the credit assignment to each predator and the reciprocal of its distance to
the prey in \textit{predator-prey} (3 agents). This experiment is conducted on 1000 randomly sampled episodes.}
\label{tab:pearson coeff}
\end{table*}
As presented in Fig~\ref{Fig.ab}, our methods STAS and STAS-ML outperform baselines, with STAS soaring to an average reward of 200 and a treasure rate of 0.9 at 30k episodes. QMIX improves slightly at 10k episodes but then declines. SQDDPG improves after 60k episodes, reaching an average reward of 130 and a treasure rate of 0.65. COMA is ineffective, with minimal rewards and a treasure rate of 0. The variance of QMIX and SQDDPG is high throughout training. Our results demonstrate our method's effectiveness in a challenging, extremely delayed reward environment, successfully learning cooperation patterns.

\subsection{Multi-agent Particle Environment}


\subsubsection{Environment settings}

We selected two scenarios from MPE, namely \textit{cooperative navigation} and \textit{predator-prey}, to evaluate the performance of all algorithms. In task \textit{cooperative navigation}, $N$ agents are asked to reach $L$ landmarks. Agents are rewarded based on how far any agent is from each landmark and penalized if they collide with other agents. 
In task \textit{predator-prey}, $N$ agents cooperate to catch $M$ preys. Agents are rewarded when any of them catch the prey and penalized if they stay away from the preys. All preys are controlled by a pre-trained policy.
In our settings, rewards are only revealed at the end of each episode.
\subsubsection{Results}
The results in Fig~\ref{Fig:MPE} demonstrate our approach's superior performance and sample efficiency over all baselines, particularly in environments with delayed rewards. Baselines show increased variance and ineffectiveness as agent numbers rise, likely due to the complexity in assessing individual agent contributions in larger cooperative systems without mid-term global rewards. While STAS-ML surpasses STAS in the \textit{predator-prey} task with three agents, it shows comparable results in other tasks. This implies that forgoing Shapley Value properties may reduce performance in complex scenarios but enhance sample efficiency in simpler, smaller-agent tasks.
\subsection{Ablations}

We conducted several ablation experiments to evaluate different components of the STAS. Fig~\ref{fig:sample_num} demonstrates the impact of different practical methods of the spatial transformer. It shows that when the sample number of Shapley value approximation is 1, not only is the performance the worst but the variance is also larger, making it less stable. When the sample number is 5, the performance is optimal, as increasing the sample number leads to a more accurate estimation of the true value. However, when the sample number is 3, the performance is already close to that of 5, indicating that the performance is good enough with 3 samples.

Fig~\ref{fig:layer_num} demonstrates the impact of using different numbers of layers in the STAS-ML. When the number of layers is 1, the performance shows higher variance and is less stable. This may be because a single layer is not sufficient to extract high-level hidden information. As the number of layers increases, the performance significantly improves. When the number of layers is 3 and 5, the performance is already very close, indicating that the model is able to extract the relevant information about the interactions between the agents with several layers.

Following \citet{DBLP:conf/aaai/WangZKG20}, we conducted an experiment to verify the effectiveness of the Shapley value. We computed the Pearson coefficient correlating the credits assigned to each agent with the inverse of their distance to the prey in a \textit{predator-prey} scenario involving three agents. Table \ref{tab:pearson coeff} indicates that our approach surpasses all baseline methods in fairness. Moreover, SQDDPG demonstrates considerable effectiveness. These results robustly confirm our rationale for integrating Shapley value to effectively address credit assignment challenges.
\section{Conclusion}
In this paper, we focus on the credit assignment problem in multi-agent learning with delayed episodic return. We introduce return decomposition methods to this problem, and we propose an novel approach called STAS.
Our approach utilizes the Shapley value to evaluate each agent's contribution and combines these contributions on both spatial and temporal scales to fit with the real episodic return from the environment. Our experimental results demonstrate that STAS outperforms all baselines, showing lower variance and faster convergence speed. In future work, we plan to include more baselines while extending it to more challenging tasks.
\section{Acknowledgments}
This work is sponsored by National Natural Science Foundation of China (62376013), Collective Intelligence \& Collaboration Laboratory (Open Fund Project No. QXZ23014101)
\bibliography{aaai24}

\end{document}